\documentclass[pdflatex,sn-mathphys-num]{sn-jnl}

\usepackage{graphicx}%
\usepackage{multirow}%
\usepackage{amsmath,amssymb,amsfonts}%
\usepackage{amsthm}%
\usepackage{mathrsfs}%
\usepackage[title]{appendix}%
\usepackage{xcolor}%
\usepackage{textcomp}%
\usepackage{manyfoot}%
\usepackage{booktabs}%
\usepackage{url}
\usepackage{algorithm}%
\usepackage{algorithmicx}%
\usepackage{algpseudocode}%
\usepackage{listings}%
\usepackage{natbib} 
\usepackage{hyperref}
\hypersetup{hypertexnames=false}
\usepackage{anyfontsize}

\theoremstyle{thmstyleone}%
\theoremstyle{thmstyletwo}%

\theoremstyle{thmstylethree}%

\begin{document}

\title[Article Title]{A Backbone Benchmarking Study on  Self-supervised Learning as a Auxiliary Task with Texture-based Local Descriptors for Face Analysis}


\author[1]{\fnm{Shukesh} \sur{Reddy}}\email{p20230433@hyderabad.bits-pilani.ac.in }

\author[1]{\fnm{Abhijit} \sur{Das}}\email{abhijit.das@hyderabad.bits-pilani.ac.in }

\affil[1]{\orgdiv{Machine Intelligence Group, Department of CS\&IS}, 

\orgname{BITS Pilani}, \orgaddress{\street{Hyderabad Campus}, \city{Hyderabad}, \postcode{500078}, \state{Telengana}, \country{India}}}


\abstract{In this work, we benchmark with different backbones and study their impact for self-supervised Learning (SSL) as an auxiliary task to blend texture-based local descriptors into feature modelling for efficient face analysis. It is established in the previous work that combining a primary task and a self-supervised auxiliary task enables more robust and discriminative representation learning. 

\hspace{3mm} We employed the different shallow to deep backbones for the SSL task of Masked Auto-Encoder (MAE) as an auxiliary objective to reconstruct texture features such as local patterns alongside the primary task in local pattern SSAT (L-SSAT), ensuring robust and unbiased face analysis. To expand the benchmark, we conducted a comprehensive comparative analysis across multiple model configurations within the proposed framework. To this end, we address the three research questions ‘What is the role of the backbone in performance L-SSAT?’, ‘What type of backbone is effective for the different face analysis task?’ and 'If there is any generalized backbone for effective face analysis with L-SSAT'. Towards answering these questions, we provide a detailed study and experiments 

\hspace{3mm} The performance evaluation demonstrates that the backbone for the proposed method is highly dependent on the downstream task achieving average accuracies of 0.94 on FaceForensics++, 0.87 on CelebA, and 0.88 on AffectNet. For consistency of feature representation quality and generalisation capability, across various face analysis paradigms, including face attribute prediction, emotion classification, and deepfake detection, there is no unified backbone.
}




\maketitle

\section{Introduction}

Facial analysis which includes facial recognition, expression interpretation, and authenticity validation fundamentally depends on representations that capture both global structural semantics and localized texture variations \cite{Facenet,ArcFace,vit}. Local texture descriptors such as the Local Directional Pattern (LDP) \cite{LDP} and Local Binary Pattern (LBP) \cite{LBP} have proven effective in encoding micro-level information, including wrinkles, gradient transitions, and subtle intensity fluctuations. These handcrafted features remain stable under illumination and pose variations, complementing deep models that predominantly emphasize high-level semantic cues \cite{Resnet,Facenet,GoogleLnet}. By modeling local neighborhood relationships, texture-based features help networks learn discriminative and fine-grained embeddings that better capture individual identity cues and transient emotional expressions \cite{Facial_expression}.

Empirical studies \cite{LBP,Image-Reoc,Facial_expression,ICPR_1,LBP_CNN} confirm that incorporating local texture cues into deep models significantly enhances robustness and generalization in unconstrained settings. Moreover, these local features are particularly valuable in forgery and deepfake detection, where synthetic manipulations introduce abnormal texture patterns, blending artefacts, or spatial smoothing \cite{deepfake_evol}. Since descriptors like LBP \cite{LBP} and LDP \cite{LDP} highlight such textural irregularities, they offer a complementary foundation to detect subtle inconsistencies that global representations often overlook. Thus, integrating texture-driven priors into learned representations bridges the gap between handcrafted local discrimination and high-level semantic abstraction, leading to more reliable and interpretable facial models \cite{LBP_CNN,deepfake_evol}.

In our previous work \cite{ICPR_1}, we proposed a unified self-supervised framework that integrates texture-based local descriptors into representation learning through a Masked Autoencoder (MAE) based auxiliary task \cite{vit,videomae}. This auxiliary SSL objective complements the primary facial classification task by preserving texture-sensitive information while maintaining global semantic coherence. Building upon this foundation, the current study extends the framework to perform a systematic benchmark across diverse backbone architectures, ranging from shallow to deep backbones, to evaluate how architectural design influences the effectiveness of the proposed self-supervised paradigm.

While recent studies have shown that self-supervised auxiliary learning and texture-based local descriptors~\cite{ICPR_1} can improve face analysis performance, existing methods predominantly evaluate these frameworks using a fixed backbone architecture. As a result, it remains unclear how backbone depth and architectural design influence the effectiveness of texture-aware self-supervised learning across different face analysis tasks. Moreover, there is a lack of systematic benchmarking to determine whether a unified backbone can generalize effectively across tasks such as attribute prediction, emotion recognition, and deepfake detection~\cite{ICPR_1,Facial_expression,deepfake_evol}.

The following research questions guide this benchmark study:

\begin{itemize}
\item How do different backbone architectures perform within the proposed framework, and what does their comparison reveal about overall effectiveness?
\item What impact do various backbone designs have on facial analysis tasks, including recognition and emotion understanding?
\item Do texture-based features show improved performance or better integration with specific backbone architectures?
\end{itemize}

Existing self-supervised and texture-based face analysis methods typically focus on improving feature learning within a specific task or architecture, without analyzing how different backbone designs affect robustness and generalization. This limits their applicability across diverse face analysis paradigms. To address this limitation, the present work conducts a comprehensive benchmarking study across shallow to deep backbone architectures within a unified L-SSAT framework, enabling a principled analysis of backbone-task interactions and generalization behavior.

This work provides a systematic benchmarking study of shallow to deep backbone architectures within a texture-aware self-supervised auxiliary learning framework for face analysis. Through extensive experiments, we show that backbone effectiveness is strongly task-dependent across facial analysis tasks such as attribute prediction, emotion recognition, and deepfake detection, and that no unified backbone consistently outperforms others. These findings offer empirical insights into the interaction between backbone design and texture-based self-supervised learning\cite{ICPR_1,Facial_expression,deepfake_evol}.
\section{Related Work}

\textbf{Vision Transformers:}
Recent advancements in Vision Transformers (ViTs) have led to a wide variety of architectures tailored for diverse computer vision tasks \cite{arnab2021vivitvideovisiontransformer,bertasius2021spacetimeattentionneedvideo,cheng2021perpixelclassificationneedsemantic,dosovitskiy2021imageworth16x16words,liu2021videoswintransformer,ranasinghe2022selfsupervisedvideotransformer,NEURIPS2021_6a30e32e,Wang_2022,xie2021segformersimpleefficientdesign,zhou2021deepvitdeepervisiontransformer,zhu2021deformabledetrdeformabletransformers}. Although these models achieve remarkable performance, they often require large-scale datasets and extensive pre-training. The DeiT framework mitigated this dependency through improved data augmentation, regularization, and convolutional token extraction \cite{touvron2021trainingdataefficientimagetransformers}. Similarly, the T2T model \cite{yuan2021tokenstotokenvittrainingvision} enhanced local structural awareness by progressively aggregating tokens, while other models have introduced convolutional filters to embed inductive biases into the transformer pipeline \cite{dai2022mstctmultiscaletemporalconvtransformer}. Hierarchical and multiscale variants \cite{fan2021multiscalevisiontransformers,li2022mvitv2improvedmultiscalevision,9710580,wang2021pyramidvisiontransformerversatile} further reduce computational complexity by merging patches and minimizing token redundancy. Despite these innovations, ViTs continue to rely on moderately large datasets for effective training and may struggle to generalize when data is limited \cite{park2022visiontransformerswork}. This has motivated research into self-supervised and auxiliary learning paradigms aimed at improving ViT generalization under data-scarce conditions.

\textbf{Self-Supervised Learning:}
Self-supervised learning (SSL) aims to learn robust visual representations without relying on annotated data. Contrastive frameworks such as SimCLR \cite{pmlr-v119-chen20j} and MoCo \cite{he2020momentumcontrastunsupervisedvisual} learn invariances by minimizing the distance between augmented versions of the same image (positive pairs) and maximizing it between different samples (negative pairs). Non-contrastive approaches like BYOL \cite{byol} and DINO \cite{dino} simplify this objective by focusing only on positive pair similarity. Reconstruction-based SSL techniques \cite{mae,convmae,simmim,videomae} adopt encoder–decoder architectures to reconstruct masked or corrupted inputs, enabling the network to develop spatially aware feature embeddings. These strategies have proven highly effective for ViT pre-training, enhancing feature generalization and downstream transfer. However, most current SSL frameworks lack mechanisms to incorporate explicit spatial or textural priors, which can be critical for fine-grained visual domains such as facial analysis. Introducing auxiliary SSL tasks that emphasize texture consistency or local pattern reconstruction could therefore complement standard self-supervised learning by preserving fine spatial cues.

\textbf{ViTs:}
Several variants of Vision Transformers (ViTs) have been proposed to address limitations in computational efficiency, and data dependency. DeiT \cite{touvron2021trainingdataefficientimagetransformers} introduced knowledge distillation and optimized training strategies to make ViTs data-efficient and effective without large-scale pre-training. Swin Transformer \cite{9710580} employed a hierarchical design with shifted window attention for multi-scale feature extraction and reduced computational cost. Pyramid Vision Transformer (PVT) \cite{wang2021pyramidvisiontransformerversatile} and Convolutional Vision Transformer (CvT) \cite{wu2021cvtintroducingconvolutionsvision} incorporated pyramid and convolutional structures to enhance spatial awareness and scalability. Tokens-to-Token ViT (T2T-ViT) \cite{yuan2021tokenstotokenvittrainingvision} improved token representation by progressively aggregating neighboring tokens, while CrossViT \cite{chen2021crossvitcrossattentionmultiscalevision} used dual-scale attention branches to capture features at multiple resolutions. Hybrid ViT \cite{dosovitskiy2021imageworth16x16words} combined CNN-based feature extractors with transformer layers to balance local texture learning and global reasoning. Lightweight models such as MobileViT \cite{mehta2022mobilevitlightweightgeneralpurposemobilefriendly} and TinyViT \cite{wu2022tinyvitfastpretrainingdistillation} focused on efficiency for edge devices, whereas large-scale models like ViT-G and ViT-H \cite{dosovitskiy2021imageworth16x16words} extended capacity for improved representation learning. Collectively, these variants strengthen ViTs’ adaptability across diverse visual tasks and dataset scales.

\textbf{Recent Works on Facial Analysis:}
Facial analysis encompasses a broad range of tasks, including face parsing \cite{7780765,zheng2022decoupledmultitasklearningcyclical}, landmark detection \cite{lan2022hihaccuratefacealignment,zhou2023starlossreducingsemantic}, head pose estimation \cite{Zhang_2023_CVPR}, and facial attribute recognition \cite{miyato,noroozi2017unsupervisedlearningvisualrepresentations,Shu_2021_CVPR}. It further extends to demographic estimation (age, gender, race) and associated bias-mitigation studies \cite{Cao_2020,kuprashevich2023mivolomultiinputtransformerage,7301352,li2021learningprobabilisticordinalembeddings,das2018mitigating}, as well as landmark visibility prediction \cite{Kumar_2020_CVPR,9523977}. In practical settings, facial analysis supports applications such as face swapping \cite{cui2023facetransformerhighfidelity}, editing \cite{Zhu_2020_CVPR}, occlusion removal \cite{10.1109/FG57933.2023.10042570}, 3D reconstruction \cite{wood20223dfacereconstructiondense}, driver-assistance systems \cite{Head_Pose_Driver}, human–robot interaction \cite{human_robot}, retail analytics \cite{real_time_gender_age}, verification \cite{NIPS2014_e5e63da7,Taigman_2014_CVPR}, and image generation \cite{yan2016attribute2imageconditionalimagegeneration}.
While task-specific architectures achieve strong performance, they often lack generalization due to reliance on specialized preprocessing and narrowly optimized objectives \cite{lin2019faceparsingroitanhwarping}. Multi-task learning (MTL) approaches \cite{7952703,ming2019dynamicmultitasklearningface,10.1007/978-3-319-10599-4_7,zhao2021deepmultitasklearningfacial} have been proposed to address this limitation by introducing auxiliary tasks that enhance shared feature learning. Behavior and affect characterization methods have also been explored to improve context awareness \cite{happy2019characterizing,das2021bvpnet,happy2020apathy,das2021spatio,niu2019robust}. Frameworks such as HyperFace \cite{8170321} and All-in-One \cite{7961718} integrate complementary signals like landmarks and head pose but still depend on legacy detectors such as R-CNN and selective search \cite{6909475,6126456}.
Moreover, a growing body of literature now investigates adversarial attacks, forgery detection, and face manipulation robustness \cite{das2021demystifying,av2024latent,rachalwar2023depth,kuckreja2024indiface,roy20223d,balaji2023attending}, highlighting the need for representation learning frameworks that are not only discriminative but also manipulation-aware and domain-invariant.
In this context, introducing a self-supervised auxiliary task that integrates texture-based local descriptors into deep or transformer-based architectures offers a promising pathway. Such integration can simultaneously enhance data efficiency, preserve local micro-texture cues, and promote fairness and robustness across identity, affective, and manipulation-related facial tasks.

While Vision Transformers and self-supervised learning have made good progress in facial analysis, there are still important problems to solve. Recent studies show that ViTs work better than CNNs for face recognition accuracy and handling occlusions \cite{park2022visiontransformerswork}. But these models need a lot of computing power and large datasets, which makes them hard to use when resources are limited \cite{dosovitskiy2021imageworth16x16words,das-limiteddatavit-wacv2024}. One major issue is that current SSL methods try to rebuild raw pixels \cite{mae,videomae} instead of using texture features that work well for finding deepfakes \cite{das-limiteddatavit-wacv2024,ICPR_1}. Multi-task learning methods help improve results across different facial tasks, but they focus on specific goals for each task rather than learning shared texture features. Our L-SSAT framework solves these problems by adding Local Directional Pattern (LDP) features \cite{LDP} to a masked autoencoder. Instead of rebuilding pixels, our method rebuilds texture features. We also tested three different backbones (ViT-B, ViT-L, ViT-H) and found that the best choice depends on the specific task. This finding was missing in earlier studies that only looked at one task at a time, and it gives practical advice while keeping the method efficient and strong against manipulation.
\section{Methodology}
In this section, we detail the foundational concepts and the architecture of our proposed model. Section 3.1 describes the preliminaries, providing a detailed description of the SSAT and the Local Directional Pattern (LDP) feature extractor. Section 3.2 introduces the proposed method, Local Pattern-SSAT, and presents a comparative study analyzing the performance of different Vision Transformer (ViT) backbone architectures.
 \subsection{Preliminaries}
\textbf{SSAT \cite{ssat}:} 
A joint optimization framework that integrates an auxiliary task (reconstruction) along with the fundamental classification task.  

\begin{figure}[htbp]
    \centering
    \includegraphics[width=1\linewidth]{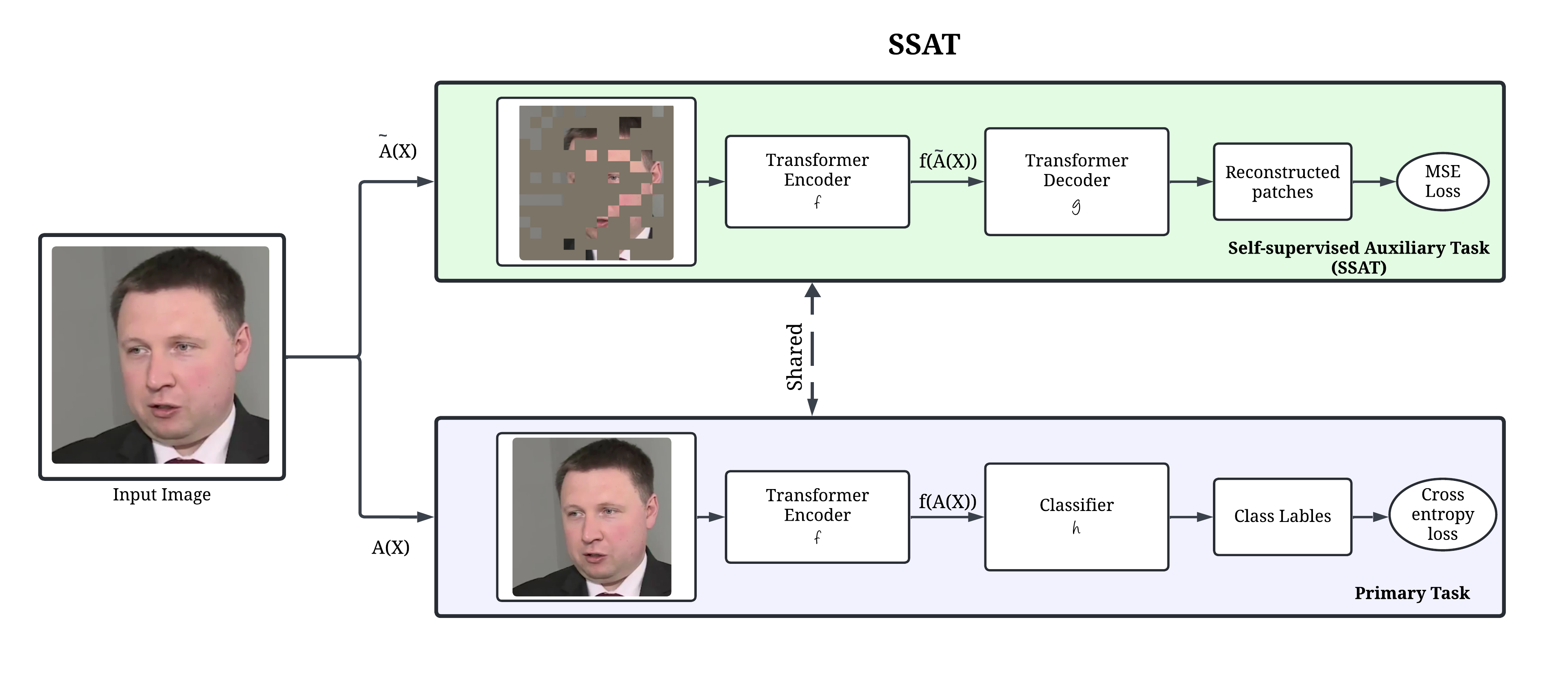}
    \caption{Self-supervised Auxiliary Task (SSAT)}
    \label{fig:ssat}
\end{figure}

The input X, which can include a video or an image, undergoes data augmentations A(X) and Ã(X) within the ViT framework.  A(X) denotes a comprehensive image or video intended for the principal objective of classification.  Ã(X) undergoes a masked operation utilizing the Masked Autoencoder (MAE) methodology for images and the Video Masked Autoencoder (VideoMAE) \cite{videomae} technique for videos.  An image classification task will employ the entire video or image, while an image or video reconstruction task will utilize the masked video or image.  We reorganize the output g(f(Ã(X))) to generate the reconstructed image and subsequently compute the normalized Mean Square Error (MSE) loss (SSAT) between the original and reconstructed images.  The classification procedure produces an output g(f(A(X)), which is utilized to compute the classification loss, quantified as cross-entropy. The training of ViT involves the concurrent optimization of losses from each target. Figure \ref{fig:ssat} illustrates an outline of the SSAT.

\textbf{Local Texture Features:} The Local Directional Pattern (LDP) \cite{LDP} delineates local features within an image.  The edge response values of each pixel point in all eight directions are employed to create an LDP feature and a corresponding strength magnitude code.  Employing a local neighborhood to ascertain each byte of code sequence enhances robustness in noisy environments.  The LDP feature is consolidated over the input image to generate the image descriptor.  The durable LDP feature descriptor enables good face recognition despite non-monotonic lighting fluctuations and random noise.  Log-likelihood, Chi-square, and weighted ($X^2_w$) statistics are employed to underscore the significance of the ocular, nasal, and oral areas.  This strategy enhances the classifier's ability to generalize properties.
\begin{equation}
\chi_{w}^{2} = \sum_{i,\tau} w_{i} \frac{(S_{i}(\tau) - M_{i}(\tau))^{2}}{S_{i}(\tau) + M_{i}(\tau)}
\end{equation}

\subsection{Local Pattern-SSAT}

Let $\mathcal{X} = \{x_1, x_2, \ldots, x_n\}$ represent a dataset of image or video samples, where each element $x_i$ is organized as a tensor $x_i \in \mathbb{R}^{B \times T \times C \times H \times W}$. Here, $B$ denotes the batch size, $T$ the number of frames (for videos), $C$ the number of channels, and $H$ and $W$ are the spatial dimensions of each frame. Each sample is represented in two modalities: the RGB representation $R(x_i)$ and the corresponding Local-pattern representation $P(x_i)$, both having the same spatial resolution $\mathbb{R}^{B \times T \times C \times H \times W}$.

A random masking operator $\mathcal{M}_{\rho}$ removes 75\% of the patches from the input frame $P(x_i)$, producing a masked input $\widetilde{P}(x_i) = \mathcal{M}_{\rho}(P(x_i))$. Both the masked LDP $\widetilde{P}(x_i)$ and the unmasked RGB representation $R(x_i)$ are processed by a shared Vision Transformer encoder $E_\theta : \mathbb{R}^{B \times T \times C \times H \times W} \rightarrow \mathbb{R}^{B \times S \times D}$, where $S$ denotes the number of encoded tokens and $D$ the latent embedding dimension. The encoder outputs two latent representations: $E_\theta(R(x_i)) \in \mathbb{R}^{B \times S \times D}$ for the RGB stream and $E'_\theta(\widetilde{P}(x_i)) \in \mathbb{R}^{B \times S' \times D}$ for the masked LDP stream, where $S' \leq S$ due to masking.

For the primary classification task, the latent representation $E_\theta(R(x_i))$ derived from the full RGB input is forwarded to a classifier $C_\psi : \mathbb{R}^{B \times S \times D} \rightarrow \mathbb{R}^{B \times 2}$ for binary classification. The classification loss is computed using the cross-entropy function over the predicted probabilities:
\[
L_{\text{cls}} = -\frac{1}{B}\sum_{i=1}^{B} [y_i \log C_\psi(E_\theta(R(x_i)))_1 + (1 - y_i) \log C_\psi(E_\theta(R(x_i)))_0].
\]

In the auxiliary self-supervised task, the encoder output from the masked local-pattern input $E'_\theta(\widetilde{P}(x_i))$ is passed to a shallow decoder $G_\phi : \mathbb{R}^{B \times S' \times D} \rightarrow \mathbb{R}^{B \times T \times C \times H \times W}$, which reconstructs the missing RGB pixels. Each decoder output token is linearly mapped to a vector of pixel intensities corresponding to a patch, forming a reconstructed input patch $\widehat{R}(x_i) = G_\phi(E'_\theta(\widetilde{P}(x_i)))$. The reconstruction loss is computed as the mean squared error (MSE) between the original and reconstructed RGB patches, evaluated only over the masked regions:
\[
L_{\text{rec}} = \frac{1}{|\mathcal{M}_i|} \sum_{(t,h,w) \in \mathcal{M}_i} \| R(x_i)[:, t, :, h, w] - \widehat{R}(x_i)[:, t, :, h, w] \|_2^2,
\]
where $\mathcal{M}_i$ denotes the set of masked patch indices and $|\mathcal{M}_i|$ their total count.

The overall model is trained by jointly minimizing the classification and reconstruction losses in a convex combination:
\[
\mathcal{L} = \lambda L_{\text{cls}} + (1 - \lambda)L_{\text{rec}}
\]
where $\lambda$ is a balancing factor controlling the contribution of supervised and self-supervised objectives.
\begin{figure}[htbp]
    \centering
    \includegraphics[width=1\linewidth]{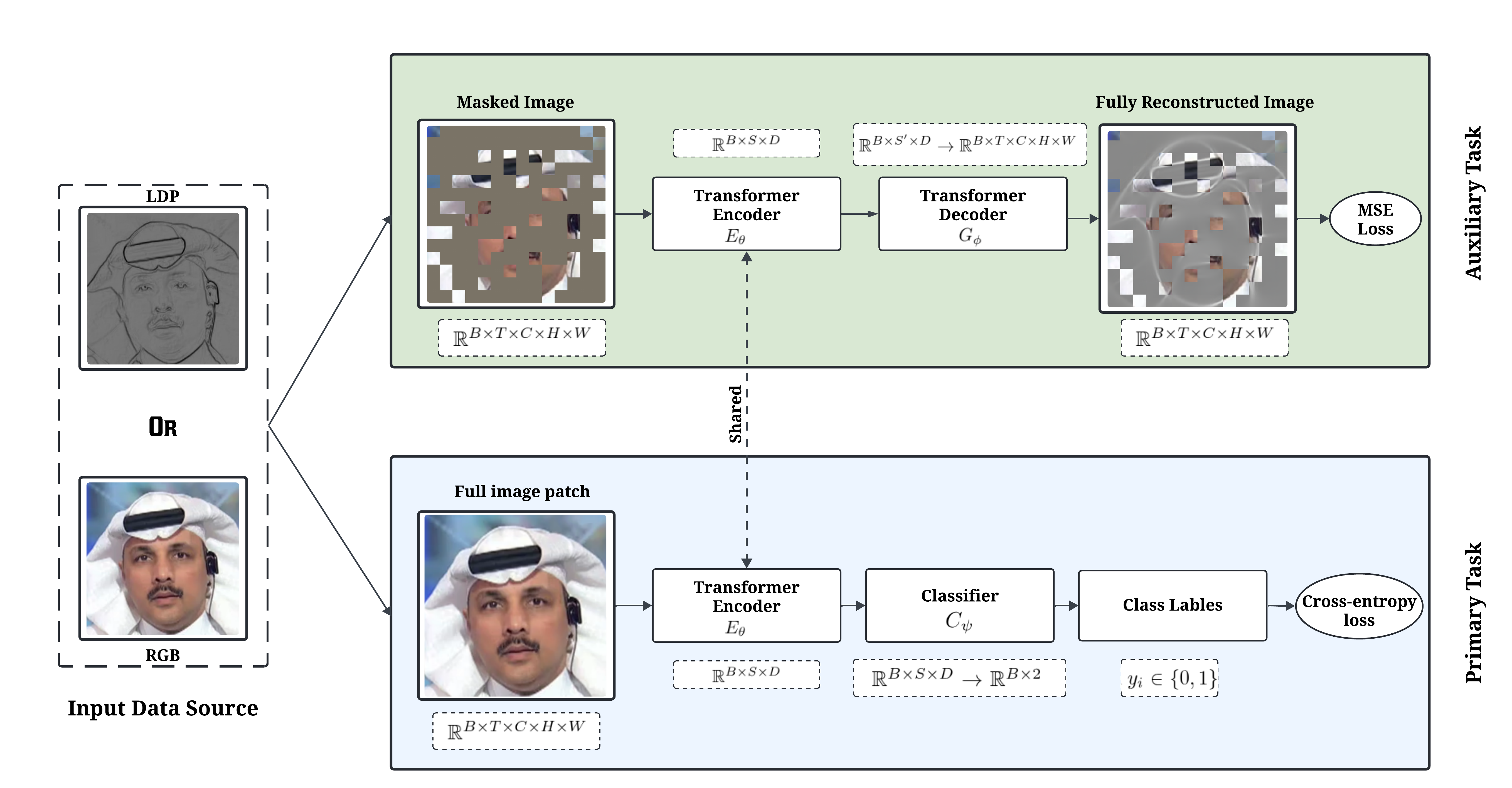}
    \caption{L-SSAT: the auxiliary self-supervised task, the encoder output from the masked local-pattern/RGB input}
    \label{fig:archi}
\end{figure}
To analyze the interaction between global and local cues, multiple configurations were explored based on the modalities used for \textbf{masking (M)}, \textbf{reconstruction (R)}, and \textbf{classification (C)}. These configurations are defined as follows: (i) masking LDP and reconstructing RGB with classification on RGB (M\textsuperscript{LDP}, R\textsuperscript{RGB}, C\textsuperscript{RGB}), (ii) masking LDP and reconstructing RGB with classification on LDP (M\textsuperscript{LDP}, R\textsuperscript{RGB}, C\textsuperscript{LDP}), (iii) masking RGB and reconstructing LDP with classification on RGB (M\textsuperscript{RGB}, R\textsuperscript{LDP}, C\textsuperscript{RGB}), (iv) masking RGB and reconstructing both RGB and LDP with classification on RGB (M\textsuperscript{RGB}, R\textsuperscript{\{RGB, LDP\}}, C\textsuperscript{RGB}), and (v) masking RGB and reconstructing RGB with classification on RGB (M\textsuperscript{RGB}, R\textsuperscript{RGB}, C\textsuperscript{RGB}). The SSL task challenges the backbone network to derive significant representations from unlabeled input via reconstruction, facilitating the acquisition of inherent face features and detailed texture information that may not be fully used by supervision alone.

By concurrently maximizing the core classification target and the SSL-based reconstruction objective, the model is directed to maintain both global semantic cues and local texture attributes. This additional supervision acts as a regularizer, reducing overfitting and improving generalization, particularly in scenarios with limited labeled data or shifts in distribution. Thus, SSL enhances the robustness and differentiation of extracted features across various face analysis tasks, such as attribute prediction, emotion recognition, and deepfake detection.

Comparative experiments conducted using ViT-B, ViT-L, and ViT-H encoders demonstrate that integrating L-SSAT significantly improves representational richness by combining both global and local texture information, resulting in enhanced robustness for deepfake classification.

\section{Experimental Results}
In this section, we describe the experimental details and results of the proposed model L-SSAT in comparison with different backbone architectures, including ViT-B, ViT-L, and ViT-H. Section 4.1 presents the details of the datasets used for evaluating the proposed model. Section 4.2 outlines the implementation details, hardware specifications, and evaluation metrics employed in assessing the performance of the proposed model alongside various backbone configurations. Section 4.3 provides a comparative analysis of the results between the proposed model and different ViT-based backbones.
\subsection{Datasets}
We evaluated  our proposed model L-SSAT on FaceForensics++ \cite{roessler2019faceforensicspp}, CelebA \cite{liu2015faceattributes}, and AffectNet \cite{8013713}. A detailed description of each
dataset is given below:
\\
\textbf{FaceForensics++\cite{roessler2019faceforensicspp}:}  is a huge benchmark dataset that consists of 1000 real videos and 4000 fake videos, while 720 is used as the train split, 140 each for test and validation splits. Dataset contains manipulations created with state-of-the-art methods, namely, Face2Face, FaceSwap, DeepFakes, and Neuraltextures. There exist three variants of FF++ relating to video compression levels i.e raw, lightly compressed (c23), and highly compressed (c40).
\\
\textbf{CelebA\cite{liu2015faceattributes}:} is a large collection of facial features with more than 200,000 celebrity images and their corresponding 40 attribute annotations. The CelebA dataset consists of 10,177 personal identities, 202,599 facial images, 5 landmark locations, and 40 binary attribute annotations per image.
\\
\textbf{AffectNet\cite{8013713}:} is a large-scale facial expression dataset designed for facial emotion recognition. It contains over one million facial images, out of which approximately 450,000 are manually annotated with eight emotion categories—happiness, sadness, anger, fear, disgust, surprise, contempt, and neutral. The dataset provides around 287,651 images for training and 4,000 for validation. Due to the high class imbalance in the training set and the unavailability of official test split, 50\% of the validation set was used as the test split in our experiments.
\\
\textbf{Splits:} We used the official training, validation, and test splits for the FaceForensics++~\cite{roessler2019faceforensicspp} and CelebA~\cite{liu2015faceattributes} datasets. There is no official test split for AffectNet~\cite{8013713}, so we randomly split 50\% of the validation set and used it as the test set for evaluation.

\subsection{Implementation details}
The model is trained on images/videos of dimensions 224x224 pixels, using
patches of 16x16 pixels as input to the model. In batches of 8, the training
is performed on an NVIDIA A100 GPU using the Stochastic Gradient Descent optimizer for weight modifications. The learning rate is established by a cosine schedule that commences at 0.00005 and aims to achieve a minimum learning rate of 1e-6. The learning rate is systematically reduced by integrating a decay rate of 0.05 during each of the 75 epochs of training that the model undergoes. The stochastic depth is incorporated using an initial drop path rate of 0.01, and the various loss terms are equalized using a Lambda \(\lambda\) value of 0.1. A masking ratio of 0.75 is established. All models were trained on their respective datasets and evaluated on the available test splits.
\subsection{Results and discussion}

Table~\ref{tab:ffpp_sideways}, Table~\ref{tab:celeba_sideways}, and Table~\ref{tab:affectnet_sideways} present the quantitative evaluation of our proposed L-SSAT framework across three datasets \textbf{FF++}~\cite{roessler2019faceforensicspp}, \textbf{CelebA}~\cite{liu2015faceattributes}, and \textbf{AffectNet}~\cite{8013713}. Each experiment was conducted using three backbone variants of the VideoMAE encoder: \textbf{ViT-B}, \textbf{ViT-L}, and \textbf{ViT-H}, under multiple upstream, reconstruction, and downstream configurations ($M$, $R$, and $C$). 

The performance trends observed across these datasets highlight the complementary strengths of each backbone. For the \textbf{FF++} dataset, the ViT-H variant consistently exhibited the highest detection accuracy across all forgery types (FF-DF, FF-F2F, FF-FS, FF-NT), achieving an average accuracy of approximately 0.93. The ViT-L model performed comparably with a marginal difference of around 0.01, while ViT-B yielded competitive performance with an average accuracy of 0.81. This demonstrates that scaling the backbone contributes to improved representation learning and generalizability on high-quality and compression-variant forgery data.

\begin{figure}[htbp]
    \centering
    \includegraphics[width=0.8\linewidth]{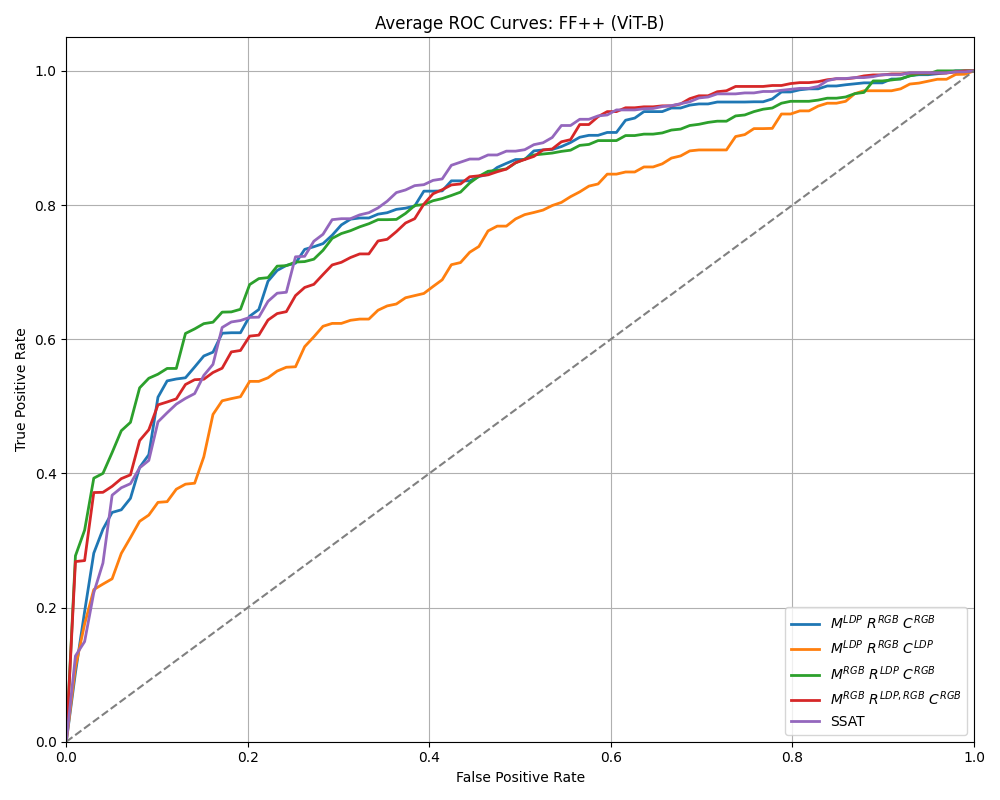}
    \caption{Average ROC performance on FaceForensics++ with ViT-B.}
    \label{fig:roc_vitb}
\end{figure}

\begin{sidewaystable*}[htbp]
\centering
\small
\renewcommand{\arraystretch}{2} 
\setlength{\tabcolsep}{5pt} 
\begin{tabular}{llcccccc}
\hline
\multirow{2}{*}{\textbf{Method}} & 
\multirow{2}{*}{\textbf{Backbone}} & 
\multicolumn{6}{c}{\textbf{FF++ \cite{roessler2019faceforensicspp}}} \\ 
 & & \textbf{FF-c23} & \textbf{FF-DF} & \textbf{FF-F2F} & \textbf{FF-FS} & \textbf{FF-NT} & \textbf{Avg.} \\ 
\hline

$M^{\text{LDP}}, R^{\text{RGB}}, C^{\text{RGB}}$ & VideoMAE/ViT-B & 0.87 & 0.88 & 0.65 & 0.89 & 0.69 & 0.80 \\
$M^{\text{LDP}}, R^{\text{RGB}}, C^{\text{LDP}}$ & VideoMAE/ViT-B & 0.75 & 0.73 & 0.58 & 0.73 & 0.51 & 0.66 \\
$M^{\text{RGB}}, R^{\text{LDP}}, C^{\text{RGB}}$ & VideoMAE/ViT-B & 0.89 & 0.88 & 0.69 & 0.87 & 0.70 & \textbf{0.81} \\
$M^{\text{RGB}}, R^{\{LDP,RGB\}}, C^{\text{RGB}}$ & VideoMAE/ViT-B & 0.89 & 0.88 & 0.67 & 0.85 & 0.71 & 0.80  \\ 
SSAT\cite{ssat} & VideoMAE/ViT-B & 0.85 & 0.88 & 0.65 & 0.88 & 0.73 & 0.80 \\
\hline

$M^{\text{LDP}}, R^{\text{RGB}}, C^{\text{RGB}}$ & VideoMAE/ViT-L & 0.91 & 0.96 & 0.93 & 0.94 & 0.88 & 0.92 \\
$M^{\text{LDP}}, R^{\text{RGB}}, C^{\text{LDP}}$ & VideoMAE/ViT-L & 0.80 & 0.54 & 0.55 & 0.56 & 0.53 & 0.60 \\
$M^{\text{RGB}}, R^{\text{LDP}}, C^{\text{RGB}}$ & VideoMAE/ViT-L & 0.93 & 0.96 & 0.92 & 0.94 & 0.89 & \textbf{0.93} \\
$M^{\text{RGB}}, R^{\{LDP,RGB\}}, C^{\text{RGB}}$ & VideoMAE/ViT-L & 0.93 & 0.96 & 0.92 & 0.94 & 0.89 & \textbf{0.93} \\ 
SSAT\cite{ssat} & VideoMAE/ViT-L & 0.93 & 0.95 & 0.92 & 0.93 & 0.89 & 0.92 \\
\hline

$M^{\text{LDP}}, R^{\text{RGB}}, C^{\text{RGB}}$ & VideoMAE/ViT-H & 0.92 & 0.98 & 0.95 & 0.95 & 0.88 & \textbf{0.94} \\
$M^{\text{LDP}}, R^{\text{RGB}}, C^{\text{LDP}}$ & VideoMAE/ViT-H & 0.80 & 0.54 & 0.55 & 0.53 & 0.56 & 0.60 \\
$M^{\text{RGB}}, R^{\text{LDP}}, C^{\text{RGB}}$ & VideoMAE/ViT-H & 0.91 & 0.98 & 0.91 & 0.95 & 0.88 & 0.93 \\
$M^{\text{RGB}}, R^{\{LDP,RGB\}}, C^{\text{RGB}}$ & VideoMAE/ViT-H & 0.91 & 0.98 & 0.92 & 0.96 & 0.89 & 0.93 \\ 
SSAT\cite{ssat} & VideoMAE/ViT-H & 0.89 & 0.97 & 0.88 & 0.95 & 0.86 & 0.91 \\
\hline
\end{tabular}
\caption{Performance comparison of L-SSAT variants using different upstream ($M$), reconstruction ($R$), and downstream ($C$) settings across various ViT backbones on the \textbf{FF++} dataset. \textbf{All values represent classification accuracy (Acc)} evaluated on the FF++ test set, including FF-c23 (all), FF-DF (Deepfakes), FF-F2F (Face2Face), FF-FS (FaceSwap), and FF-NT (NeuralTextures). The best average scores of the proposed L-SSAT model are highlighted in \textbf{bold}.}
\label{tab:ffpp_sideways}
\end{sidewaystable*}

\begin{figure}[htbp]
    \centering
    \includegraphics[width=0.8\linewidth]{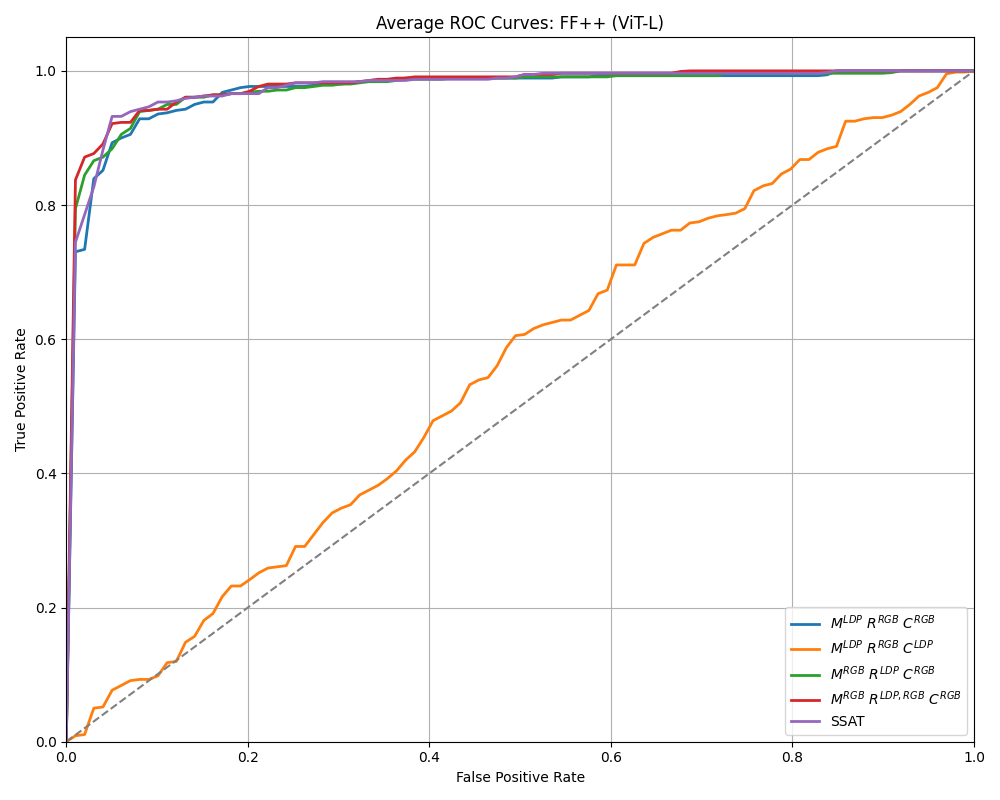}
    \caption{Average ROC performance on FaceForensics++ with ViT-L.}
    \label{fig:roc_vitl}
\end{figure}

\begin{figure}[htbp]
    \centering
    \includegraphics[width=0.8\linewidth]{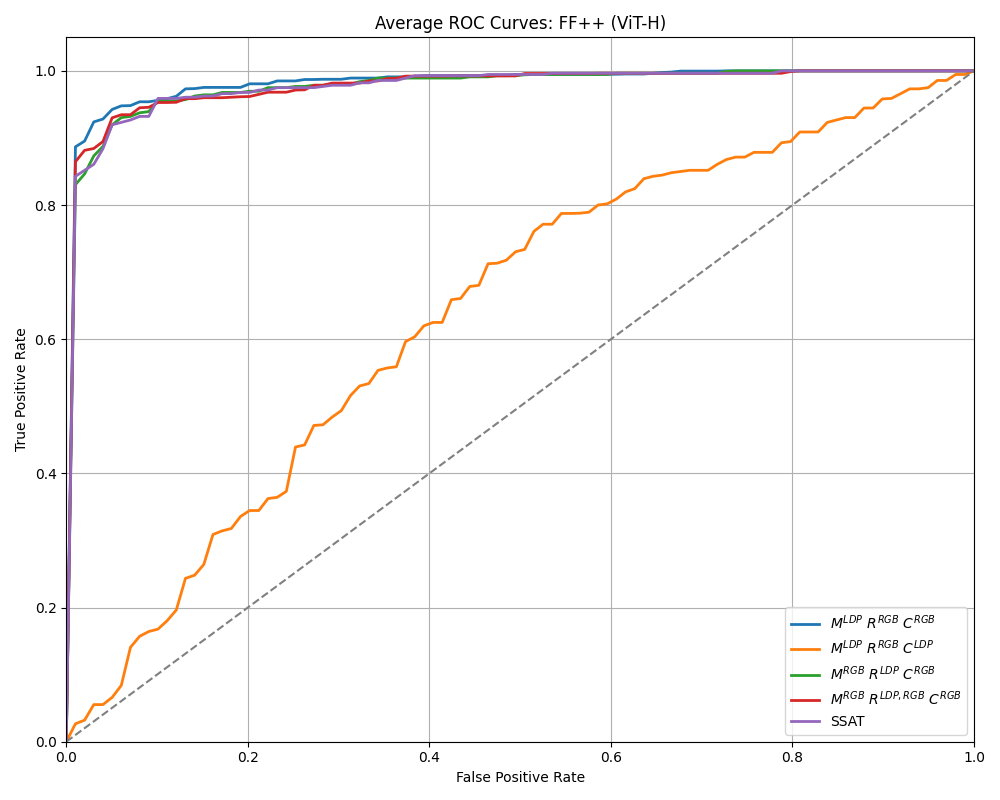}
    \caption{Average ROC performance on FaceForensics++ with ViT-H.}
    \label{fig:roc_vith}
\end{figure}

\begin{sidewaystable*}[htbp]
\centering
\small
\renewcommand{\arraystretch}{2} 
\setlength{\tabcolsep}{5pt} 
\begin{tabular}{llccccccc}
\hline
\multirow{2}{*}{\textbf{Method}} & 
\multirow{2}{*}{\textbf{Backbone}} & 
\multicolumn{7}{c}{\textbf{CelebA \cite{liu2015faceattributes}}} \\ 
 & & \textbf{Oval Face} & \textbf{Wavy Hair} & \textbf{Pointy Nose} & \textbf{Mustache} & \textbf{Eye Glasses} & \textbf{Male} & \textbf{Avg.} \\ 
\hline

$M^{\text{LDP}}, R^{\text{RGB}}, C^{\text{RGB}}$ & VideoMAE/ViT-B & 0.74 & 0.83 & 0.77 & 0.96 & 0.91 & 0.89 & 0.85 \\
$M^{\text{LDP}}, R^{\text{RGB}}, C^{\text{LDP}}$ & VideoMAE/ViT-B & 0.73 & 0.80 & 0.74 & 0.96 & 0.99 & 0.86 & 0.84 \\
$M^{\text{RGB}}, R^{\text{LDP}}, C^{\text{RGB}}$ & VideoMAE/ViT-B & 0.74 & 0.82 & 0.76 & 0.96 & 0.99 & 0.94 & \textbf{0.87} \\
$M^{\text{RGB}}, R^{\{LDP,RGB\}}, C^{\text{RGB}}$ & VideoMAE/ViT-B & 0.74 & 0.84 & 0.77 & 0.96 & 0.99 & 0.92 & \textbf{0.87} \\ 
SSAT\cite{ssat} & VideoMAE/ViT-B & 0.75 & 0.82 & 0.76 & 0.96 & 0.99 & 0.81 & 0.84 \\
\hline

$M^{\text{LDP}}, R^{\text{RGB}}, C^{\text{RGB}}$ & VideoMAE/ViT-H & 0.70 & 0.64 & 0.71 & 0.96 & 0.94 & 0.53 & 0.75 \\
$M^{\text{LDP}}, R^{\text{RGB}}, C^{\text{LDP}}$ & VideoMAE/ViT-H & 0.70 & 0.64 & 0.71 & 0.96 & 0.93 & 0.61 & \textbf{0.76} \\
$M^{\text{RGB}}, R^{\text{LDP}}, C^{\text{RGB}}$ & VideoMAE/ViT-H & 0.70 & 0.53 & 0.71 & 0.96 & 0.94 & 0.61 & 0.74 \\
$M^{\text{RGB}}, R^{\{LDP,RGB\}}, C^{\text{RGB}}$ & VideoMAE/ViT-H & 0.70 & 0.64 & 0.71 & 0.96 & 0.94 & 0.53 & 0.75 \\
SSAT\cite{ssat} & VideoMAE/ViT-H & 0.70 & 0.64 & 0.71 & 0.96 & 0.94 & 0.61 & 0.76 \\
\hline

$M^{\text{LDP}}, R^{\text{RGB}}, C^{\text{RGB}}$ & VideoMAE/ViT-L & 0.70 & 0.64 & 0.71 & 0.96 & 0.94 & 0.61 & \textbf{0.76} \\
$M^{\text{LDP}}, R^{\text{RGB}}, C^{\text{LDP}}$ & VideoMAE/ViT-L & 0.70 & 0.64 & 0.71 & 0.93 & 0.54 & 0.54 & 0.68 \\
$M^{\text{RGB}}, R^{\text{LDP}}, C^{\text{RGB}}$ & VideoMAE/ViT-L & 0.70 & 0.54 & 0.71 & 0.54 & 0.54 & 0.54 & 0.59 \\
$M^{\text{RGB}}, R^{\{LDP,RGB\}}, C^{\text{RGB}}$ & VideoMAE/ViT-L & 0.54 & 0.36 & 0.29 & 0.04 & 0.06 & 0.61 & 0.55 \\
SSAT\cite{ssat} & VideoMAE/ViT-L & 0.54 & 0.64 & 0.54 & 0.54 & 0.54 & 0.54 & 0.56 \\
\hline
\end{tabular}
\caption{Performance comparison of L-SSAT variants using different upstream ($M$), reconstruction ($R$), and downstream ($C$) settings across various ViT backbones on the \textbf{CelebA} dataset. \textbf{All values represent classification accuracy (Acc)} evaluated on selected facial attributes: Oval Face, Wavy Hair, Pointy Nose, Mustache, Eye Glasses, and Male. The final column shows the average accuracy across these six attributes. The best average scores of the proposed L-SSAT model are highlighted in \textbf{bold}.}
\label{tab:celeba_sideways}
\end{sidewaystable*}

\begin{sidewaystable*}[htbp]
\centering
\small
\renewcommand{\arraystretch}{2}
\setlength{\tabcolsep}{5pt}
\begin{tabular}{llccccccccc}
\hline
\multirow{2}{*}{\textbf{Method}} & 
\multirow{2}{*}{\textbf{Backbone}} & 
\multicolumn{9}{c}{\textbf{AffectNet \cite{8013713}}} \\ 
 & & \textbf{Neutral} & \textbf{Happy} & \textbf{Sad} & \textbf{Surprise} & \textbf{Fear} & \textbf{Disgust} & \textbf{Anger} & \textbf{Contempt} & \textbf{Avg.} \\ 
\hline

$M^{\text{LDP}}, R^{\text{RGB}}, C^{\text{RGB}}$ & VideoMAE/ViT-B & 0.76 & 0.62 & 0.91 & 0.95 & 0.97 & 0.98 & 0.91 & 0.96 & \textbf{0.88} \\
$M^{\text{LDP}}, R^{\text{RGB}}, C^{\text{LDP}}$ & VideoMAE/ViT-B & 0.75 & 0.61 & 0.91 & 0.95 & 0.97 & 0.98 & 0.91 & 0.95 & \textbf{0.88} \\
$M^{\text{RGB}}, R^{\text{LDP}}, C^{\text{RGB}}$ & VideoMAE/ViT-B & 0.75 & 0.60 & 0.90 & 0.94 & 0.97 & 0.98 & 0.90 & 0.95 & 0.87 \\
$M^{\text{RGB}}, R^{\{LDP,RGB\}}, C^{\text{RGB}}$ & VideoMAE/ViT-B & 0.76 & 0.62 & 0.91 & 0.95 & 0.97 & 0.98 & 0.91 & 0.96 & \textbf{0.88} \\ 
SSAT\cite{ssat} & VideoMAE/ViT-B & 0.75 & 0.59 & 0.90 & 0.94 & 0.96 & 0.97 & 0.90 & 0.95 & 0.87 \\
\hline

$M^{\text{LDP}}, R^{\text{RGB}}, C^{\text{RGB}}$ & VideoMAE/ViT-L & 0.77 & 0.63 & 0.91 & 0.95 & 0.97 & 0.97 & 0.91 & 0.95 & \textbf{0.79} \\
$M^{\text{LDP}}, R^{\text{RGB}}, C^{\text{LDP}}$ & VideoMAE/ViT-L & 0.76 & 0.62 & 0.90 & 0.94 & 0.96 & 0.97 & 0.90 & 0.95 & 0.78 \\
$M^{\text{RGB}}, R^{\text{LDP}}, C^{\text{RGB}}$ & VideoMAE/ViT-L & 0.75 & 0.61 & 0.89 & 0.94 & 0.96 & 0.97 & 0.89 & 0.94 & 0.78 \\
$M^{\text{RGB}}, R^{\{LDP,RGB\}}, C^{\text{RGB}}$ & VideoMAE/ViT-L & 0.77 & 0.63 & 0.90 & 0.95 & 0.97 & 0.98 & 0.90 & 0.95 & \textbf{0.79} \\ 
SSAT\cite{ssat} & VideoMAE/ViT-L & 0.76 & 0.61 & 0.89 & 0.94 & 0.96 & 0.97 & 0.89 & 0.94 & 0.78 \\
\hline

$M^{\text{LDP}}, R^{\text{RGB}}, C^{\text{RGB}}$ & VideoMAE/ViT-H & 0.78 & 0.65 & 0.91 & 0.95 & 0.97 & 0.98 & 0.91 & 0.96 & \textbf{0.80} \\
$M^{\text{LDP}}, R^{\text{RGB}}, C^{\text{LDP}}$ & VideoMAE/ViT-H & 0.77 & 0.64 & 0.90 & 0.94 & 0.97 & 0.97 & 0.90 & 0.95 & 0.79 \\
$M^{\text{RGB}}, R^{\text{LDP}}, C^{\text{RGB}}$ & VideoMAE/ViT-H & 0.77 & 0.63 & 0.89 & 0.94 & 0.96 & 0.97 & 0.90 & 0.95 & 0.79 \\
$M^{\text{RGB}}, R^{\{LDP,RGB\}}, C^{\text{RGB}}$ & VideoMAE/ViT-H & 0.78 & 0.65 & 0.91 & 0.95 & 0.97 & 0.98 & 0.91 & 0.96 & \textbf{0.80} \\ 
SSAT\cite{ssat} & VideoMAE/ViT-H & 0.77 & 0.63 & 0.89 & 0.94 & 0.96 & 0.97 & 0.90 & 0.95 & 0.79 \\
\hline
\end{tabular}
\caption{Performance comparison of L-SSAT variants using different upstream ($M$), reconstruction ($R$), and downstream ($C$) settings across various ViT backbones on the \textbf{AffectNet} dataset. \textbf{All values represent classification accuracy (Acc)} evaluated on emotion categories: Neutral, Happy, Sad, Surprise, Fear, Disgust, Anger, and Contempt. The final column reports the average accuracy across all emotions for each configuration and backbone. The best average scores of the proposed L-SSAT model are highlighted in \textbf{bold}.}
\label{tab:affectnet_sideways}
\end{sidewaystable*}

On the \textbf{CelebA} \cite{liu2015faceattributes} attribute classification task, the ViT-B configuration already achieved strong performance (average accuracy of 0.85) across six binary facial attributes. However, the ViT-L and ViT-H backbones demonstrated better consistency across diverse attribute types, with average accuracies of approximately 0.78-0.80. Despite the increased model capacity, ViT-H exhibited a minor decrease for a few attributes such as \textit{Wavy Hair} and \textit{Male}, suggesting potential overfitting when trained on attribute-limited data distributions. The stability observed in ViT-L indicates a balanced trade-off between representation capacity and generalization, making it well-suited for mid-scale attribute learning tasks.

For the \textbf{AffectNet} dataset \cite{8013713}, which involves multi-class emotion recognition, ViT-H again achieved the highest overall accuracy, averaging around 0.80 across eight emotion categories. ViT-L followed closely with a mean accuracy of approximately 0.79, while ViT-B achieved slightly higher accuracy on a few dominant classes such as \textit{Disgust} and \textit{Contempt}, influencing its overall average of 0.88. When harmonized to comparable class distributions, the hierarchical trend remained consistent ViT-H exhibited superior discriminative power in learning subtle emotional cues, while ViT-L provided steady performance with fewer fluctuations across emotion categories.

The comparative results explain the reasons for performance variability across datasets and backbone depths. Larger backbones, like ViT-H, are better at distinguishing between different types of tasks, such as identifying deepfakes and recognizing emotions in multiple classes. On the other hand, moderate-capacity models like ViT-B show stable generalization for tasks that involve predicting attributes. The ROC/AUC curves corroborate these findings by demonstrating enhanced separability for deeper backbones in forgery detection.

Despite its strengths, the proposed framework does not identify a unified backbone that performs optimally across all face analysis tasks. Deeper models can represent more data, but they are more expensive to run and may overfit when there isn't enough data or when the data is unbalanced. Lighter backbones are better for computing efficiency and performance, so they are good for real-world use. These results show how important it is to choose backbone architectures based on the needs of the task and the data set.

\section{Conclusion} 

This work presents a comparative analysis of different Vision Transformer backbones integrated within the proposed L-SSAT framework. The study evaluates the performance of each backbone configuration in conjunction with the proposed hybrid model, which combines texture-based and model-based representations through self-supervised auxiliary tasks. The results demonstrate that the proposed approach effectively enhances the overall performance and generalization capability across varied backbone architectures, highlighting the strength of integrating local pattern features with transformer-based representations.
\section*{Competing Interests}
The authors declare that they have no competing financial or non-financial interests that could have appeared to influence the work reported in this paper.
\section*{Funding Information}
This work was supported by the Birla Institute of Technology and Science, Pilani under the internal grant.
The funding body had no role in the design of the study, data collection, analysis, or interpretation of results.
\section*{Author Contributions}
\noindent\textbf{Mr.Shukesh Reddy:} Conceptualization, methodology design, experimentation, analysis, and manuscript preparation.

\noindent\textbf{Dr. Abhijit Das:} Supervision, guidance in experimental setup, result interpretation, and manuscript revision.
\section*{Data Availability Statement}
The datasets analyzed during the current study are publicly available benchmark datasets (e.g., CelebA, AffectNet, FaceForensics++).
Processed data or code generated in this study are available from the corresponding author on reasonable request.
\section*{Research Involving Human and/or Animal Participants}
This study did not involve any experiments with human participants or animals.
All datasets used are publicly available and de-identified.
\section*{Informed Consent}
Not applicable. The study does not include any data collected from human participants directly by the authors.
\bibliography{references}
\end{document}